\documentclass[letterpaper, 10 pt, conference]{ieeeconf}  

\IEEEoverridecommandlockouts                              

\usepackage{graphicx} 
\usepackage{amsmath} 
\usepackage{amssymb}  
\usepackage{multirow}
\usepackage{subfigure}
\graphicspath{ {./images/} }

\title{\LARGE \bf
Lvio-Fusion: A Self-adaptive Multi-sensor Fusion SLAM Framework Using Actor-critic Method
}

\author{Yupeng Jia$^{1}$, Haiyong Luo$^{2}$$^{*}$, Fang Zhao$^{1}$, Guanlin Jiang$^{1}$, Yuhang Li$^{1}$, \\Jiaquan Yan$^{1}$, Zhuqing Jiang$^{3}$ and Zitian Wang$^{4}$
\thanks{$^{1}$ 
School of Computer Science (National Pilot Software Engineering School), Beijing University of Posts and Telecommunications, Beijing, China}%
\thanks{$^{2}$ Haiyong Luo is with the Beijing Key Laboratory of Mobile Computing and Pervasive Device, Institute of Computing Technology, Chinese Academy of Sciences, Beijing 100190, China}%
\thanks{$^{3}$ Zhuqing Jiang is with the School of Artificial Intelligence, Beijing University of Posts and Telecommunications, Beijing, China}%
\thanks{$^{4}$ School of Artificial Intelligence, Beijing Technology and Business University, Beijing, China}%
\thanks{$^{*}$ Corresponding author, email addresses: yhluo@ict.ac.cn}%
\thanks{This work was supported in part by the National Key Research and Development Program under Grant 2019YFC1511400, the Action Plan Project of the Beijing University of Posts and Telecommunications supported by the Fundamental Research Funds for the Central Universities under Grant 2019XD-A06, the National Natural Science Foundation of China under Grant 61872046 and 62002026, the Joint Research Fund for Beijing Natural Science Foundation and Haidian Original Innovation under Grant L192004, the Key Research and Development Project from Hebei Province under Grant 19210404D, the Science and Technology Plan Project of Inner Mongolia Autonomous Regio under Grant 2019GG328 and the Open Project of the Beijing Key Laboratory of Mobile Computing and Pervasive Device.}
}

\begin{document}

\maketitle
\thispagestyle{empty}
\pagestyle{empty}

\begin{abstract}

State estimation with sensors is essential for mobile robots. Due to different performance of sensors in different environments, how to fuse measurements of various sensors is a problem. In this paper, we propose a tightly coupled multi-sensor fusion framework, Lvio-Fusion, which fuses stereo camera, Lidar, IMU, and GPS based on the graph optimization. Especially for urban traffic scenes, we introduce a segmented global pose graph optimization with GPS and loop-closure, which can eliminate accumulated drifts. Additionally, we creatively use a actor-critic method in reinforcement learning to adaptively adjust sensors' weight. After training, actor-critic agent can provide the system better and dynamic sensors' weight. We evaluate the performance of our system on public datasets and compare it with other state-of-the-art methods, which shows that the proposed method achieves high estimation accuracy and robustness to various environments. And our implementations are open source and highly scalable.

\end{abstract}

\section{INTRODUCTION}

Mobile robots are complex systems that integrate computer technology, sensor technology, information processing, electronic engineering, automation, and artificial intelligence. With the development of technology, mobile robot with more functionality can be used in emergency rescue, industrial automation and other fields. Simultaneous Localization and Mapping (SLAM) is the fundamental technical basis for mobile robots to complete tasks autonomously. At the same time, with the continuous development of robot technology and the increasing application of robots, a single sensor cannot meet the increasingly abundant robot functional requirements, thus multi-source perception information fusion technology has gradually received widespread attention.

To achieve real-time high-precision positioning and mapping to support mobile robots' six-axis state estimation, researchers have conducted much research, including visual-based methods and lidar-based methods. But those methods have some shortcomings. On the one hand, the sensitivity of monocular or stereo camera used in vision-based methods to initialization, illumination, and range contributes to instability. On the other hand, The sparse information offered by lidar makes it difficult to further directly dig sematic information. As lidar's cost continues to decrease, more user scenarios will choose to use both vision and lidar together. However, lidar and camera will inevitably produce accumulated drifts and cannot be used in large-scale environments. Therefore, the introduction of sensors such as GPS, barometer, and magnetometer can effectively eliminate accumulated drifts and provide positioning in the global coordinate system.

In this paper, we propose a tightly coupled multi-source fusion framework Lvio-Fusion, which fuses stereo camera, Lidar, IMU and GPS. The system extracts keyframes through visual-inertial odometry and optimizes factor graphs based on keyframes with other sensors. The IMU with higher frequency provides the original motion estimation for the visual odometer and eliminates the laser point cloud's distortion. In the global pose graph optimization stage, the trajectory is divided into multiple segments according to the differences of accumulated errors generated when the vehicle is turning or not. Each segment is optimized separately. Additionally, we explore the solution of the problem that different sensors have different confidences in different environments when the system is running. With the optimization of the factor graph, we run a parrallel reinforcement learning thread to adjust the different factors’ weight. The main contributions of our work can be summarized as follows:

• A tightly coupled multi-sensor fusion framework with stereo camera, lidar, IMU, and GPS based on optimization, which achieves highly accurate, real-time mobile robot trajectory estimation.

• A segmented global optimization method for urban traffic scenes, which can eliminate accumulated drifts and provide a global position.

• A self-adaptive algorithm using actor-critic method, which can adaptively adjust sensors' weight for different environments.

\section{RELATED WORK}

In the last few decades, researchers have proposed many impressive methods based on different sensors, in which vision-based methods \cite{orbslam, DSO, SVO} and LiDAR-based methods \cite{loam, lego} earn massive attention. For sensors used in state estimation, such as IMU, their fusion design schemes can usually be grouped into loosely coupled sensor fusion and tightly coupled sensor fusion. Tightly coupled systems use raw measurements of IMU in pose estimates rather than treating IMU as an independent module. For example, \cite{loam, rtvio} can be classified as loosely coupled methods. Though loosely coupled fusion is a more straightforward way to deal with different sensors, more methods base on tightly coupled sensor fusion. And \cite{beforevinsmono, vinsmono} introduce various formulas that integrate visual and inertial data and implement more efficient tightly coupled systems.

Further on, current multi-source fusion methods can be roughly divided into filtering-based methods, optimization-based methods, and deep learning-based methods. Filtering-based methods include vision-based methods \cite{msckf, ekfvio, rubustekf, ukffusion}, and LIC-fusion\cite{licfusion, licfusion2} which includes stereo camera, laser and IMU methods. They usually use EKF (Extended Kalman Filter), UKF (Unscented Kalman Filter), or MSCKF (Multi-State Constraint Kalman Filter) for pose estimation, which are fast and need low computational cost, but are sensitive to time synchronization. Since subsequent measurements in filtering process may cause troubles, a special ordering mechanism is required to ensure that all measurements' sequence from different sensors are correct. However, the optimization-based methods \cite{vinsmono, vinsfusion, LIO-SAM, vilslam} reduce the sensibility of time synchronization by using local maps or sliding windows. Furthermore, they have the ability to optimize history poses and achieve real-time performance with the aid of bundle adjustment. Also, the optimization-based method is more flexible. Besides, methods based on deep learning are more novel. DeLS-3D\cite{DeLS-3D} is a deep learning sensor fusion method that combines camera images, IMU, GPS, and 3D semantic graphs; \cite{deep2} proposes an end-to-end learnable architecture which exploits both lidars as well as cameras to perform accurate localization; And \cite{deep3} fuses 3D lidar and monocular camera for urban road detection. However, methods based on deep learning face the challenges of applicability and performance.

\section{A OPTIMIZATION-BASED FRAMEWORK WITH MULTIPLE SENSORS}

\subsection{System Overview} 

We first define the coordinate system and symbols used throughout our paper. We denote the world frame as W and the body frame as B. For convenience,  we assume that the robot's body frame's position coincides with the IMU frame's position. The forward direction of the robot is the positive direction of the X-axis, the left for that of Y-axis, and the upward for that of Z-axis. The robot state $\mathcal{X}$ can be written as:
\begin{equation}
  \mathcal{X}=[\mathbf{R}^T,\mathbf{p}^T,\mathbf{v}^T,\mathbf{b}^T]^T\label{state}
\end{equation}
where $\mathbf{R} \in SO(3)$ is the rotation matrix, $\mathbf{p} \in \mathbb{R}^3$ is the position of the robot, $\mathbf{v}$ is the speed, and $\mathbf{b}$ is the IMU bias.

We design a framework, trying to combine sensors' information and the robot state $\mathcal{X}$, and introduce various constraints to estimate the state. This type of problem can be described as a maximum a posteriori (MAP) problem. Therefore, under the assumption of the Gaussian noise model, the MAP inference for our problem is equivalent to solving the nonlinear least-squares problem \cite{factorgraph}.

\begin{equation}
  \begin{array}{c}
    \arg \min \limits_{\mathcal{X}_{i}}
    \left\{
      \sum\limits_{i \in \mathcal{B}} w_{b_i} \left\|\mathbf{r}_{\mathcal{B}_i}\right\|^2+ 
      \sum\limits_{(i, j) \in \mathcal{C}} w_{c_i} \rho\left(\left\|\mathbf{r}_{\mathcal{C}_{i,j}}\right\|^{2}\right) \right. \\ \left. + 
      \sum\limits_{(i, l) \in \mathcal{L}} w_{l_i} \rho\left(\left\|\mathbf{r}_{\mathcal{L}_{i,l}}\right\|^{2}\right)\right\}
    \end{array}\label{least-squares}
\end{equation}

$\mathbf{r}_{\mathcal{B}_i}$, $\mathbf{r}_{\mathcal{C}_{i,j}}$ and $\mathbf{r}_{\mathcal{L}_{i,l}}$ represent inertial,
visual and lidar factors's residual respectively. $\rho( \cdot )$ represents robust huber norm \cite{huber}. $w_{b_i}$, $w_{c_i}$ and $w_{l_i}$ are factors' weight.

\begin{figure*} 
    \centering
    \includegraphics{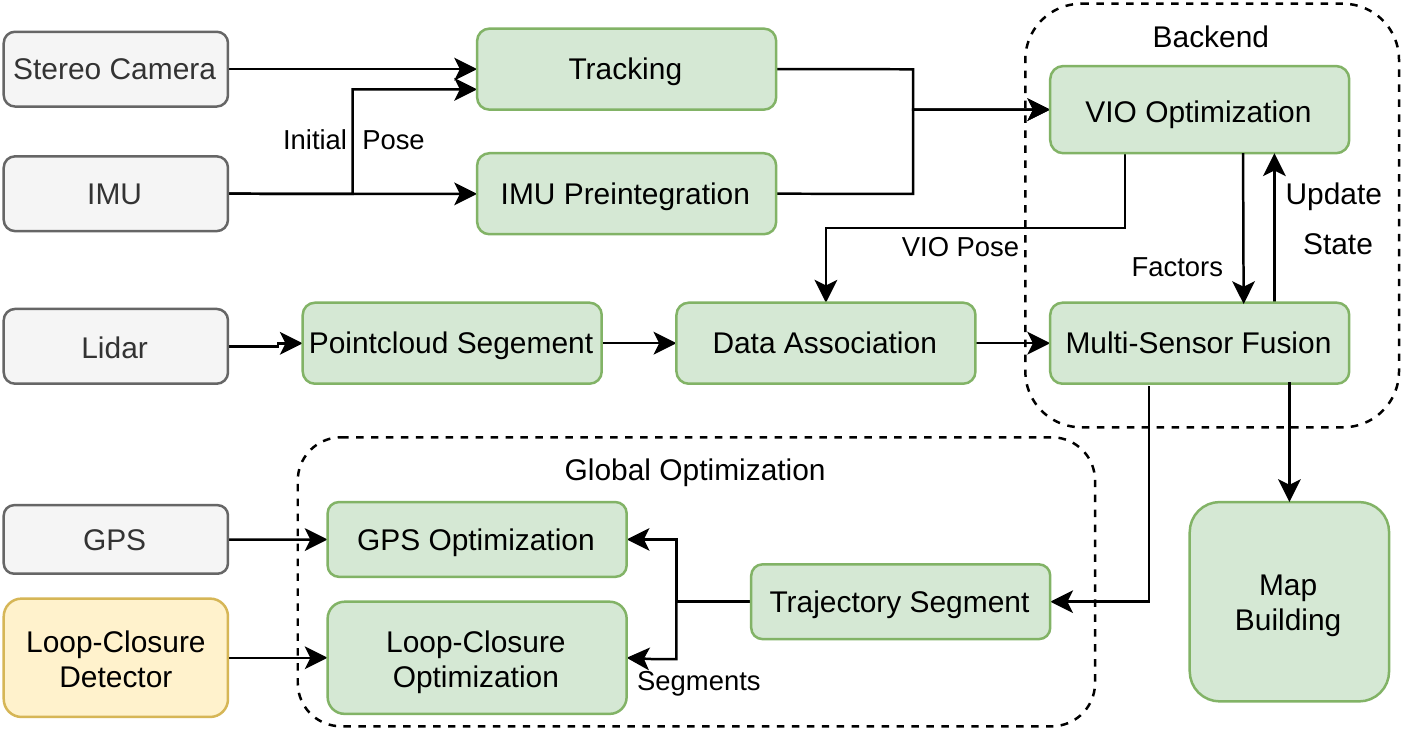}
    \caption{The system diagram of Lvio-Fusion. Arrows indicate how messages flow within the system.}
  \label{fig:overview}
\end{figure*}

An overview of our system is shown in Fig.\ref{fig:overview}. The visual frontend obtains 3D landmarks from stereo camera, performs frame-to-frame tracking and triangulates features. At the same time, high-frequency IMU measurements provide initial poses for frame-to-frame tracking, which will be calculated in IMU pre-integration. And then our system performs the VIO local optimization the first time. For lidar, our system extracts a point cloud scan centered on the visual keyframe and then segments the ground point clouds. With the pose obtained in visual-inertial odometry as the initial pose, our system uses the feature association method mentioned in \cite{loam} and performs local optimization the second time, which includes lidar, visual and IMU. In addition, we also introduce GPS constraint and loop-closure constraint to optimize the global pose graph. There are five types of factors in our system: (a) visual factors, (b) IMU pre-integration factors, (c) lidar factors, (d) GPS factors, and (e) loop closure factors. We will describe the five types of factors in the following sections.

\subsection{Visual Factors}

In visual frontend tracking,  we combine the KLT sparse optical flow algorithm \cite{klt} and the feature-based method. The KLT sparse optical flow algorithm is used for frame-to-frame tracking. This method is fast but not very stable. In the case that the robot violently shakes or illumination changes, it is easy to lose track, even if there is an IMU for motion estimation. To fix this, we build a local map with feature points of keyframes in a sliding window. When the number of tracked feature points is smaller than the threshold, the frontend will mark this frame as a keyframe and detect new ORB\cite{orb} feature points. Then, those new features which are successfully matched between the left and right camera are triangulated into the real-world position.  When every keyframe comes, the system will search old feature points in the local map to match new feature points by their descriptors.

Consider the $l^{th}$ feature that is observed in the $i^{th}$ keyframe, the residual for visual factor is defined as:
\begin{equation}
  \mathbf{r}_{\mathcal{C}_{i,j}} = \mathbf{u}_{j} -  \mathbf{K} \exp \left(\mathbf{\xi}_i^{\wedge}\right) \mathbf{P}_{j}\label{visual factor}
\end{equation}
where $\mathbf{u}_{j}$ is the feature's coordinate on the image, $\mathbf{K}$ is the camera projection matrix, and $\mathbf{P}_{j}$ is the point's position in $\mathbf{W}$. $\mathbf{\xi}_i \in SE(3)$ means the camera's pose.

\subsection{IMU Pre-integration Factors}

The raw angular velocity and acceleration measurements from an IMU are formulated as:
\begin{equation}
\hat{\mathbf{\omega}}_t=\mathbf{\omega}_t+\mathbf{b}^\omega_t+\mathbf{n}^\omega_t \label{imu measurements 1}
\end{equation}
\begin{equation}
\hat{\mathbf{a}}_t = \mathbf{R}^{BW}_t(\mathbf{a}_t-\mathbf{g})+\mathbf{b}^a_t+\mathbf{n}^a_t \label{imu measurements 2}
\end{equation}
where $\hat{\mathbf{\omega}}^t$ and $\hat{\mathbf{a}}^t$ are the raw IMU measurements in $B$ at time $t$. $\hat{\mathbf{\omega}}^t$ and $\hat{\mathbf{a}}^t$ are affected by a bias $\mathbf{b}_t$ and additive noise $\mathbf{n}_t$. We assume that the additive noise are Gaussian. Acceleration bias and gyroscope bias are modeled as random walk. $\mathbf{R}^{BW}_t$ is the rotation matrix from $W$ to $B$. $\mathbf{g}$ is the constant gravity vector in $W$.

IMU pre-integration was first proposed in \cite{preintegrate} and further improved in \cite{preintegrate2} by adding post-IMU deviation correction. We apply the IMU pre-integration residual proposed in \cite{vinsmono}, which uses IMU pre-integration to obtain the relative pose between two keyframes. The factor of IMU pre-integration can be defined as Eqs.\ref{imu factor}. 

\begin{equation}
\begin{aligned}
&\mathbf{r}_{\mathcal{B}_i}
= \left[\begin{array}{c}
  \mathbf{r}_{\mathbf{v}_i}\\
  \mathbf{r}_{\mathbf{p}_i}\\
  \mathbf{r}_{\mathbf{R}_i}\\
  \mathbf{r}_{\mathbf{b}_{a_i}}\\
  \mathbf{r}_{\mathbf{b}_{g_i}}
\end{array}\right] \\
&=
\left[\begin{array}{c}
  \mathbf{R}_{i-1}^{\top}\left(\mathbf{v}_{i}-\mathbf{v}_{i-1}-\mathbf{g} \Delta t_{i}\right) - \Delta \hat{\mathbf{v}}_{i} \\
  \mathbf{R}_{i-1}^{\top}\left(\mathbf{p}_{i}-\mathbf{p}_{i-1}-\mathbf{v}_{i-1} \Delta t_{i}-\frac{1}{2} \mathbf{g} \Delta t_{i}^{2}\right) -\Delta \hat{\mathbf{p}}_{i} \\
  \mathbf{R}_{i-1}^{\top} \mathbf{R}_{i} \hat{\mathbf{R}}_{i-1}^{i^{\top}}\\
  \mathbf{b}_{a_i}-\mathbf{b}_{a_{i-1}} \\
  \mathbf{b}_{g_i}-\mathbf{b}_{g_{i-1}}
\end{array}\right]
\end{aligned}\label{imu factor}
\end{equation}
where $[\hat{\mathbf{v}}_{i},\hat{\mathbf{p}}_{i},\hat{\mathbf{R}}_{i-1}^{i}]^{\top}$ are pre-integrated IMU measurement terms using only noisy accelerometer and gyroscope measurements between two consecutive keyframes.

In IMU pre-integration factor, initial values of speed, bias, and gravity direction are required, so IMU initialization is very important. When the system is at its beginning stage or IMU is interrupted, IMU initialization will be conducted. In IMU initialization, our system selects the most recent ten keyframes as a window and performs three steps. Firstly, the accelerometer's average value in the window is used as the initial value of the gravitational acceleration, and the speed is calculated by the prior poses. Secondly, the optimization only involving the IMU pre-integration factor is operated, and the bias velocity and gravity are estimated. And then, the final step is a full VIO optimization together with the visual factor.

\subsection{Lidar Factor}

\begin{figure} 
  \centering
  \includegraphics[width=8cm]{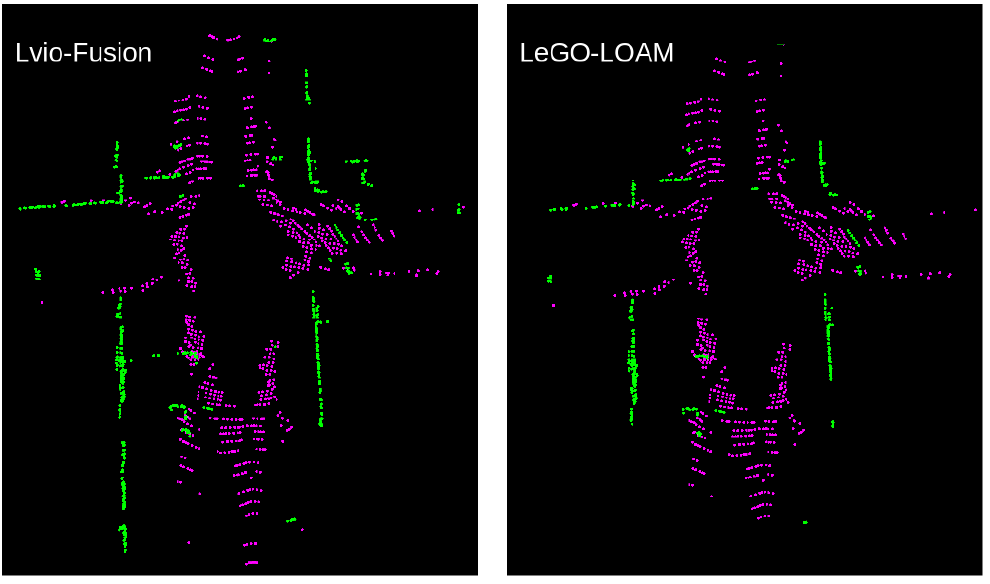}
  \caption{The image on the left is a plane point cloud extracted using the proposed curvature formula, and the right is a plane point cloud extracted by the curvature formula of lego loam.}
\label{fig:lidar}
\end{figure}

Our system generates lidar factors referring to \cite{loam}. When a new lidar scan arrives, we first extract planar features by evaluating the curvature of adjacent points. Points with small curvature values are classified as plane features. The curvature formulas in \cite{loam} and \cite{lego} ignore the angle between the laser and the plane. In our system, the Eqs.\ref{pointcloud cov 1} and Eqs.\ref{pointcloud cov 2} are used to calculate curvatures. $r_{i}$ is a adjacent point's range and $a$, $n$ are two custom constants. As shown in Fig. \ref{fig:lidar}, our system can extract the plane that extends farther, which is important since lidar scans in keyframes are not continuous.

\begin{equation}
\Delta r=(r_{i-n}-r_{i+n})/2n\label{pointcloud cov 1}
\end{equation}
\begin{equation}
c=\frac{a}{(2n-1)r_i}\displaystyle \sum^{2n-1}_{k=1}{(r_{i-n+k}-r_{i-n}-k\Delta r)^2}\label{pointcloud cov 2}
\end{equation}

Inspired by \cite{lego}, a two-step optimization for ground feature points and planar feature points segmented by the angle between laser and ground is applied. To further improve, the SAC (Sample Consensus) plane segmentation is utilized to eliminate the outliers of the ground plane. Next, our system builds a local map composed by the latest three scans. The lidar factor is defined by Eqs.\ref{lidar factor}. $j$, $u$, $v$, and $w$ are the point indices in their feature set, respectively. Only plane features are used in our system because experiments show that edge features are unstable.

\begin{equation}
  \mathbf{r}_{\mathcal{L}_{i,l}}=\frac{\bigg|
  \begin{aligned} 
  (\mathbf{p}_{i,j} & -\mathbf{p}_{l,u})\\
  (\mathbf{p}_{l,u}-\mathbf{p}_{l,v}) & \times(\mathbf{p}_{l,u}-\mathbf{p}_{l,w})
  \end{aligned} 
  \bigg|}{| (\mathbf{p}_{l,u}-\mathbf{p}_{l,v})\times(\mathbf{p}_{l,u}-\mathbf{p}_{l,w}) |}\label{lidar factor}
\end{equation}

\subsection{Pose Graph Optimization with GPS and Loop Closure}

Although we have IMU, lidar, and stereo camera used in the pose estimation, and can achieve good positioning accuracy in the short term, but accumulated errors inevitably occur. Especially for rotation of the robot, once the direction deflects a slight angle, the trajectory's error will be amplified continuously. To solve this problem, we introduce global pose graph optimization with GPS constraint and loop closure constraint, which can correct the trajectory in the long-term, large-scale map, because their error will not accumulate.

\begin{figure} 
    \centering
    \includegraphics{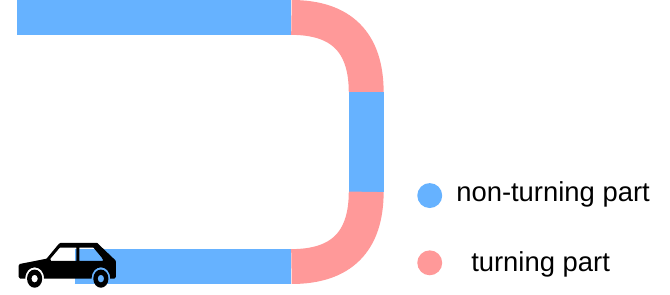}
    \caption{The illustration of the segmented pose graph.}
  \label{fig:pose graph}
\end{figure}

During real-world experiments, it is found that the rotation error of the pose estimation of the mobile vehicle when turning is much larger than that of the non-turning part. For the reason above, we creatively utilize different strategies to optimize different parts of trajectory. We segment the trajectory into multiple parts according to its steering angle, as shown in Fig.\ref{fig:pose graph}, and optimize the pose graph by parts.

In the segmented pose graph optimization, there are two stages. First, we optimize parts with GPS constraints and loop-closure constraints; Second, keyframes in the same part will be optimized, according to whether it is in a turning or a non-turning part. The rotation errors in the keyframes of turning parts are mainly optimized, while the forward direction's errors in the keyframes of non-turning parts are mainly optimized.

\section{A SELF-ADAPTIVE WEIGHTS ADJUSTMENT ALGORITHM USING ACTOR-CRITIC METHOD}

\subsection{Algorithm Framework}

In the graph based optimization with different factors, \cite{vinsfusion} uses Mahalanobis distance to express the dimensionless distance related to the distribution. But considering the complexity and diversity of environments, we design a self-adaptive method using deep reinforcement learning. The reason for using deep reinforcement learning is that pose estimation with different types of factors is too difficult to backpropagate, while deep reinforcement learning can update parameters by policy gradient algorithm. In the standard reinforcement learning model, the agent is connected to its environment through perception and action. In each step of interaction, the agent receives observations from the current state of the environment as input; the agent then chooses an action to generate as output, which will change the state of the environment. And the value of this state transition is communicated to the agent through a reward. The agent's behavior should choose actions that tend to increase the long-run sum of values of the reward.\cite{rl}

Based on the above, the environment of our framework is the graph based multi-sensor fusion optimization, the action is the weight of various factors, the observation is the information perceived by the robot from the sensors in the current state and the reward is defined as the reciprocal of RPE(Relative Pose Error) between the estimated pose and the real pose.

\begin{equation}
reward_i=\left\| \left(\mathbf{R}_{i}^{-1} \mathbf{R}_{i+\Delta}\right)^{-1}\left(\mathbf{p}_{i}^{-1} \mathbf{p}_{i+\Delta}\right)  \right\|^{-1} \label{reward}
\end{equation}

\begin{figure} 
  \centering
  \includegraphics[width=8cm]{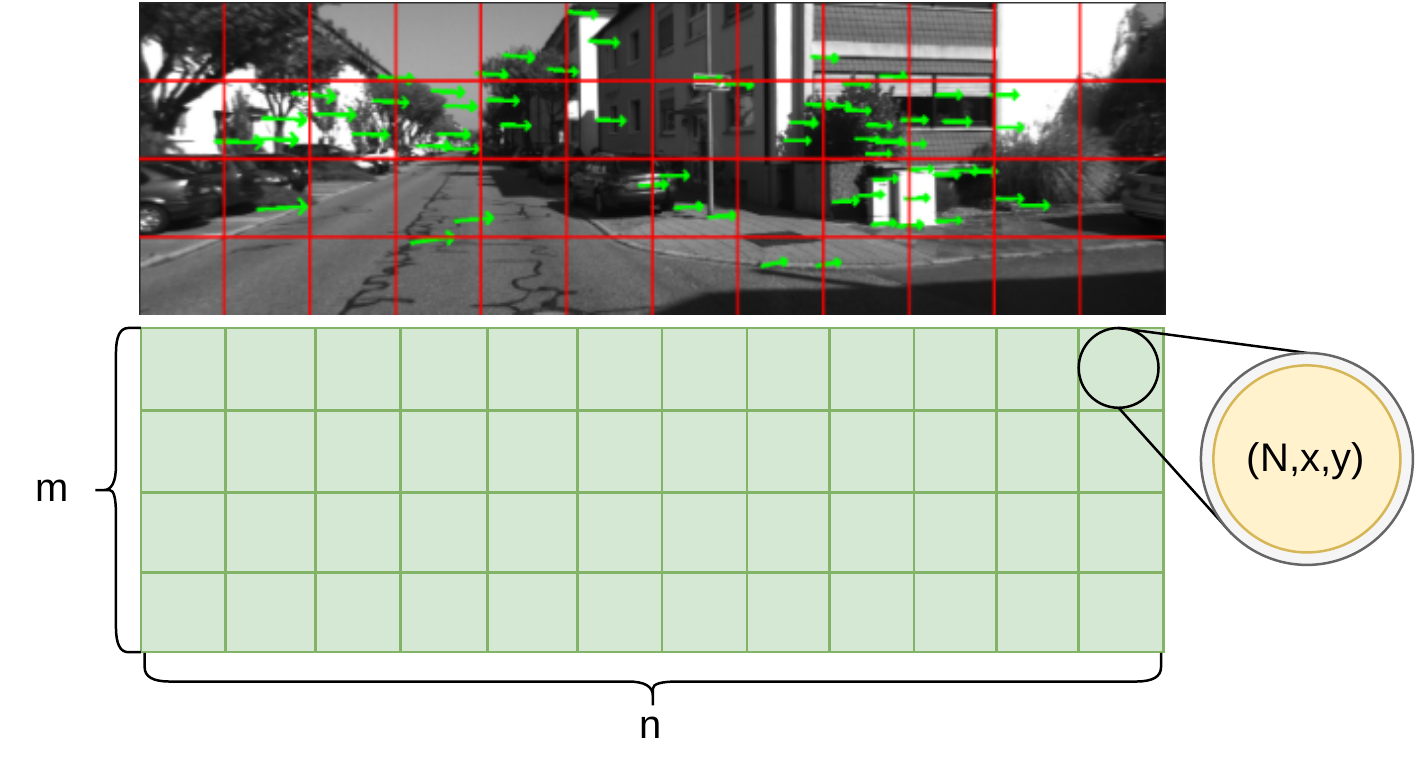}
  \caption{The data structure of observation. The image is divided into a grid of $m \times n$. N is the number of feature points in the grid; x and y denote the average movement of the matched features between this frame and the previous frame.}
\label{fig:obs}
\end{figure}

\begin{figure*}[h] 
  \centering
  \includegraphics{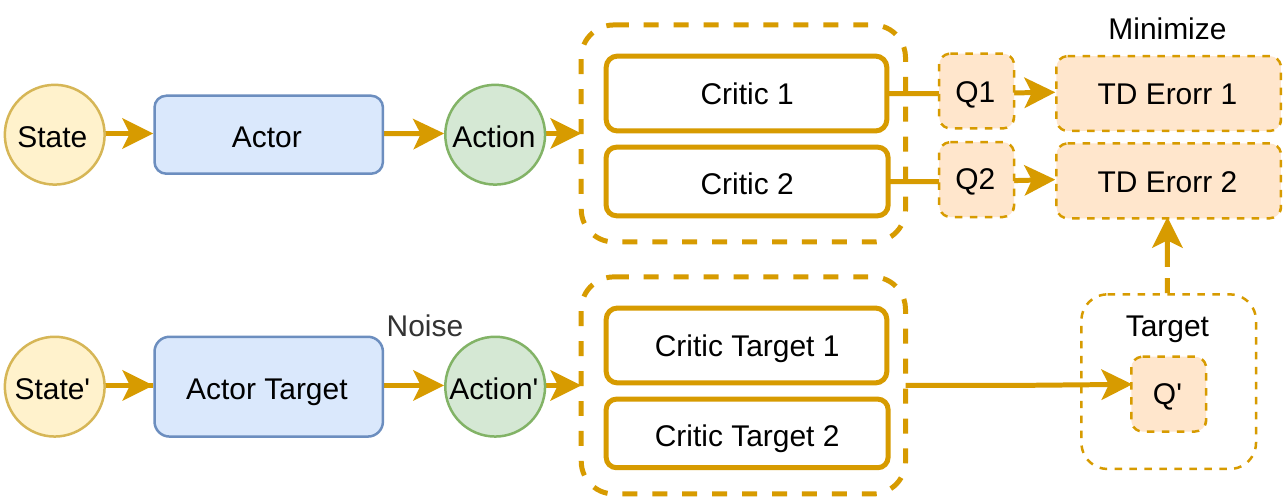}
  \caption{The network architecture of TD3.}
\label{fig:rl model}
\end{figure*}

For better results and real-time performance, we design a lightweight model using the matrix composed of feature tracking information from the visual frontend as observation instead of raw images, as shown in Fig. \ref{fig:obs}. This matrix contains motion information and environmental information. Our network takes the observation matrix as input and the weights, including the visual factor's weight and lidar factors' weight, as output. TD3 (Twin Delayed Deep Deterministic policy gradient algorithm) \cite{td3} is applied to maximize the long-term reward, which is an actor-critic reinforcement learning algorithm that computes a continuous optimal policy. We set six MLPs(Multi-Layer Perceptron) as backbones in the networks --- actor, actor target, cirtic 1, cirtic 2, cirtic target 1 and cirtic target 2, as shown in Fig.\ref{fig:rl model}. The training stage is described in Section \ref{sec:experiment-rl}.

\section{EXPERIMENT}

We evaluate the proposed system on the KITTI dataset \cite{kitti} and the KAIST Urban dataset \cite{kaist}. By comparison with other state of the art multi-sensor fusion algorithms, the accuracy and performance of the proposed system are shown. Our implementation of Lvio-Fusion is available on Github$\footnote{https://github.com/jypjypjypjyp/lvio\_fusion}$.

\subsection{Evaluation on KITTI Dataset}

KITTI dataset is collected onboard a car, which is equipped with a stereo camera(PointGray Flea2 grayscale),  a lidar (Velodyne HDL-64E), an IMU (within OXTS RT3003), and a GPS (within OXTS RT3003). The calibrations of transformation between sensors are also provided. Besides, it contains ground truth poses provided by OXTS RT3003. 

\begin{figure} 
  \centering
  \includegraphics[width=8cm]{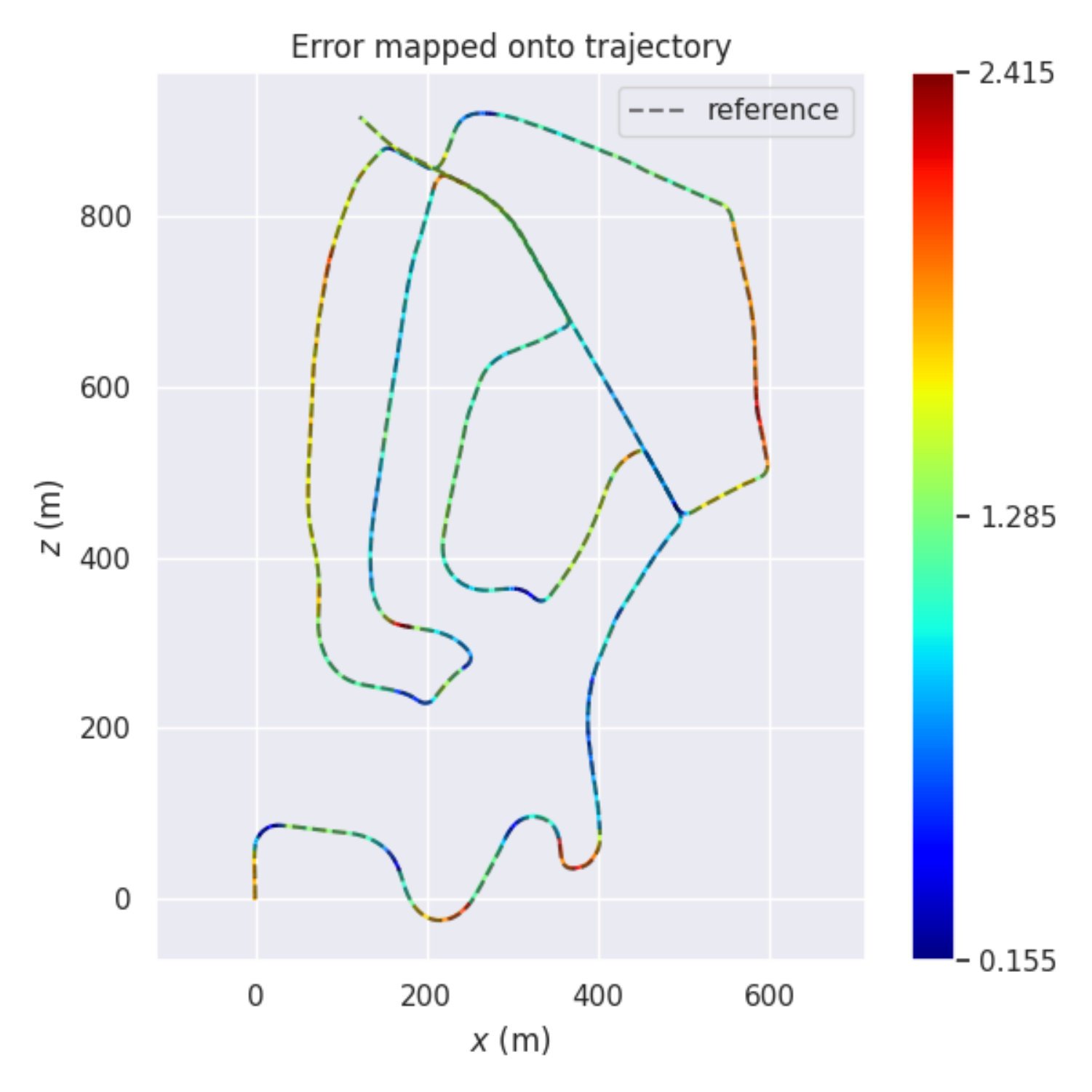}
  \caption{ATE mapped onto the trajectory result of sequence 02 output by Lvio-Fusion.}
\label{fig:lvio2}
\end{figure}

In the experiment, we used stereo camera's images, lidar's point cloud scans, and raw measurements of IMU and GPS, and compared our system with state of the art multi-sensor fusion SLAM algorithms VINS-Fusion and LIO-SAM. The former, VINS-Fusion, is a visual-inertial odometer and fuses GPS with local estimations. The latter, LIO-SAM, is a lidar-inertial odometer, which includes a global optimization with loop closure constraints and GPS constraints. All these methods are implemented using C++ and ROS and executed on a laptop with an Intel i7-7700HQ CPU (up to 3.80GHz) and 16GB memory in Ubuntu Linux. We evaluated ATE (Absolute Trajectory Error) by tool \cite{evo}. The result of sequences which are longer in size and more complex in scenarios shown in Table \ref{tab:kitti}. Fig. \ref{fig:lvio2} is the output trajectory of Lvio-Fusion on sequence 00. In this experiment, IMU and GPS's interruption occur at some moments, which causes drift using both VINS-Fusion and LIO-SAM. Our method is not dependent on a certain sensor, so interruption does not affect our result much.

Timing statistics are show in Table \ref{tab:time}. And we conducted stress tests by accelerating the dataset playback speed to evalues overall time-consuming of our algorithm. In expriments, all KITTI dataset sequences passed the 3x speed tests and retained good results as usual, which proves that outstanding performance of our method in accuracy and time overhead.

\begin{table}[h]
  \centering
  \caption{ATE in KITTI dataset in meters}
  \label{tab:kitti}
  \begin{tabular}{c|cccccc}
  \hline
  \multirow{2}{*}{Seq.} & \multicolumn{2}{c|}{VINS-Fusion} & \multicolumn{2}{c|}{LIO-SAM} & \multicolumn{2}{c}{Lvio-Fusion} \\ \cline{2-7}
  & mean   & RMSE  & mean    & RMSE       & mean       & RMSE     \\ \cline{1-7}
  00 & 1.18    & 1.24 & fail          & fail          & \textbf{1.07} & \textbf{1.13}     \\ 
  02 & 2.56    & 3.62 & 4.56          & 5.43          & \textbf{1.26} & \textbf{1.32}     \\ 
  05 & 1.19    & 1.26 & \textbf{0.99} & \textbf{1.09} & 1.13          & 1.16              \\ 
  06 & \textbf{0.97} & \textbf{1.03} & 17.69         & 12.50         & 1.06          & 1.18              \\ 
  07 & 0.88    & 0.98 & \textbf{0.37} & \textbf{0.39} & 0.73          & 0.81              \\ 
  09 & 0.83    & 0.90 & 8.03          & 9.87          & \textbf{0.82} & \textbf{0.88}     \\ \hline
  \end{tabular}%
\end{table}

\begin{table}
  \centering
  \caption{TIMING STATISTICS}
  \label{tab:time}
  \begin{tabular}{ccc}
  \hline
  \textbf{Thread} &\textbf{Module} & \textbf{Time(s)} \\ \hline
  \rule{0pt}{10pt}
    1 & Frontend tracking & 0.039 \\ \hline
  \rule{0pt}{10pt}
    2 &  Backend optimization & 0.346 \\ \hline
  \end{tabular}
\end{table}

\subsection{Evaluation on KAIST Urban Dataset}

\begin{figure} 
  \centering
  \includegraphics[width=8cm]{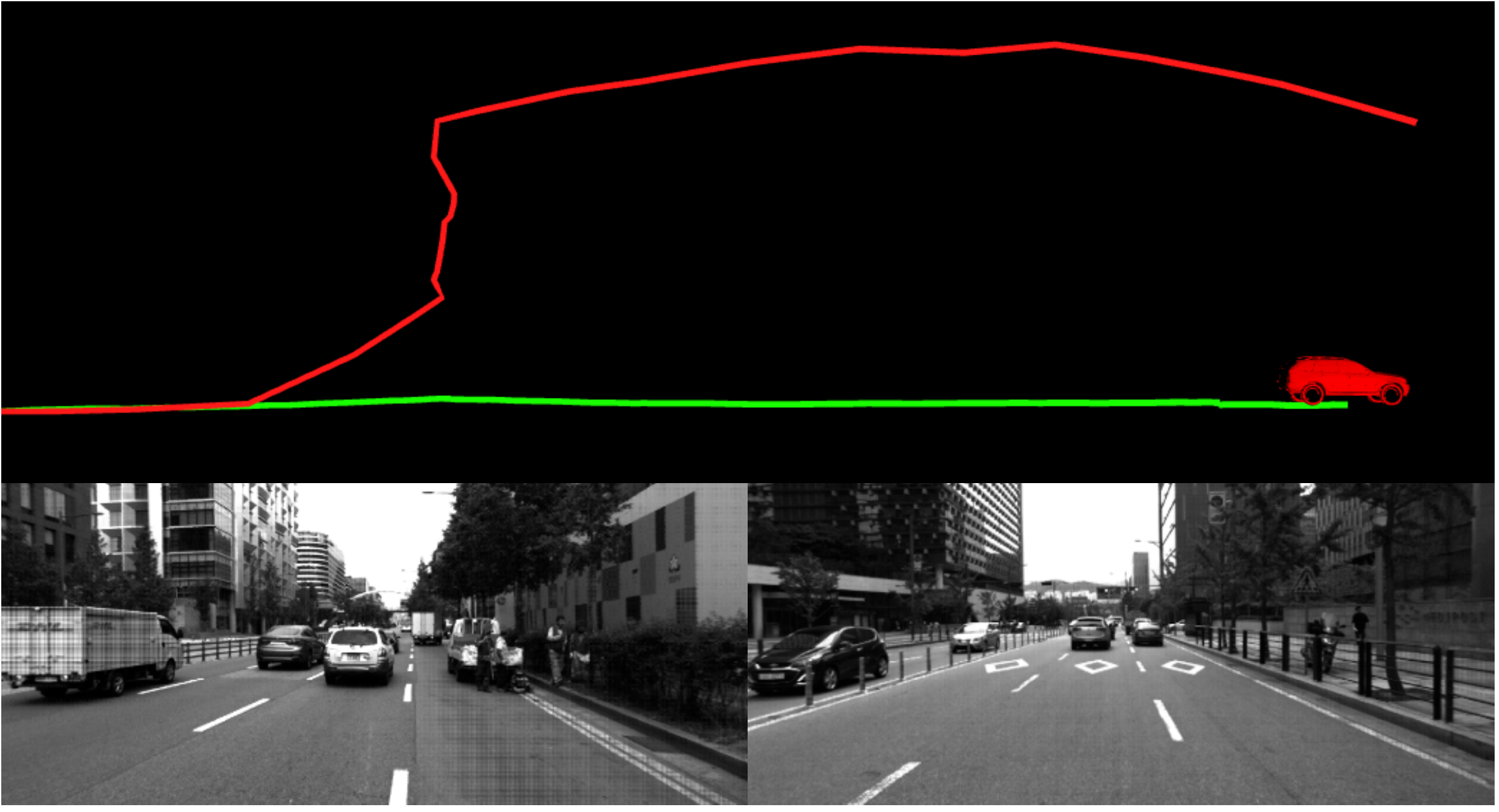}
  \caption{The bottom of this figure is two images in KAIST urban dataset. At the top of the figure, the red line shows GPS trajectory, which drifts seriously, and the green line shows the trajectory estimated by Lvio-Fusion.}
\label{fig:kaisturban}
\end{figure}

KAIST Urban dataset is a more complex dataset as shown as Fig.\ref{fig:kaisturban}. It is collected in downtown. While getting accurate positioning in large-scale scenes is already a challege, moving vehicles and pedestrians even lift it up to another difficult level. Consequently, multi-sensor fusion and global constraints are more important. The devices of KAIST Urban dataset contains two Pointgrey flea3 cameras, two Velodyne VLP-16 lidars, an Xsens MTi-300 IMU, and a U-Blox EVK-7P GPS. 

In this experiment, we did not use lidars because they are not set horizonally and compared our method with VINS-Fusion. The results of ATE are shown in Table \ref{tab:kaist}. It is worth noting that the GPS drifted greatly due to the blocking and reflection of satellites' signal from high-rise buildings. Benefiting from visual-inertial odometry and segmented global optimization, the system avoids the impact of GPS drift well with only a simple GPS filter, which is shown in Fig.7.

\begin{table}[h]
  \centering
  \caption{ATE in KAIST Urban dataset in meters}
  \label{tab:kaist}
  \begin{tabular}{c|cccc}
  \hline
  \multirow{2}{*}{Seq.} & \multicolumn{2}{c|}{VINS-Fusion} & \multicolumn{2}{c}{Lvio-Fusion} \\ \cline{2-5}
  & mean   & RMSE  & mean   & RMSE        \\ \cline{1-5}
  \rule{0pt}{10pt} 
  urban39-pankyo & 63.30    & 70.69 & \textbf{5.79} & \textbf{6.31}   \\ 
  \hline
  \end{tabular}%
\end{table}

\begin{figure}[h]
  \centering
  \subfigure[Lvio-Fusion]{
  \begin{minipage}[t]{1\linewidth}
  \centering
  \includegraphics[width=8cm]{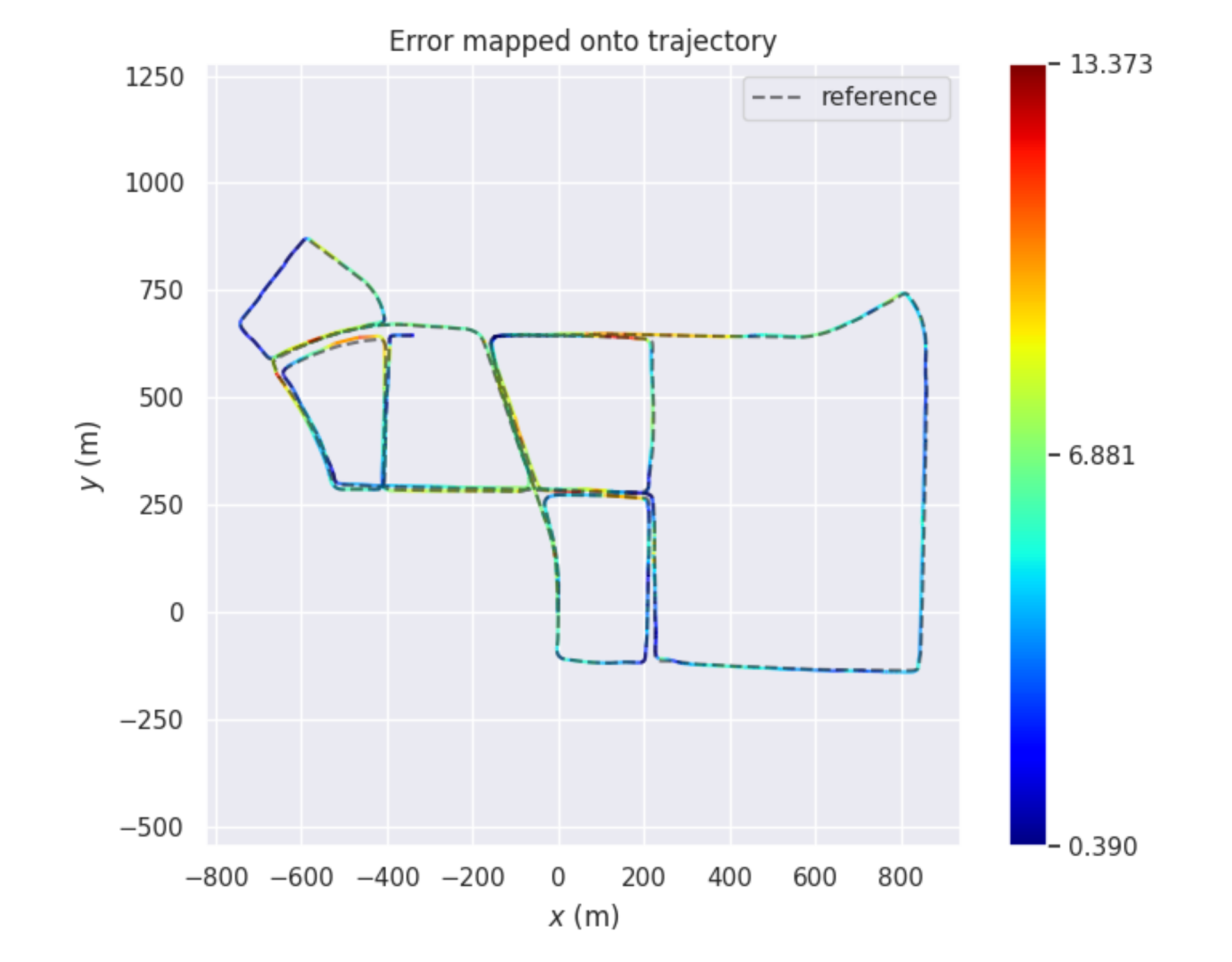}
  \end{minipage}%
  }%

  \subfigure[VINS-Fusion]{
  \begin{minipage}[t]{1\linewidth}
  \centering
  \includegraphics[width=8cm]{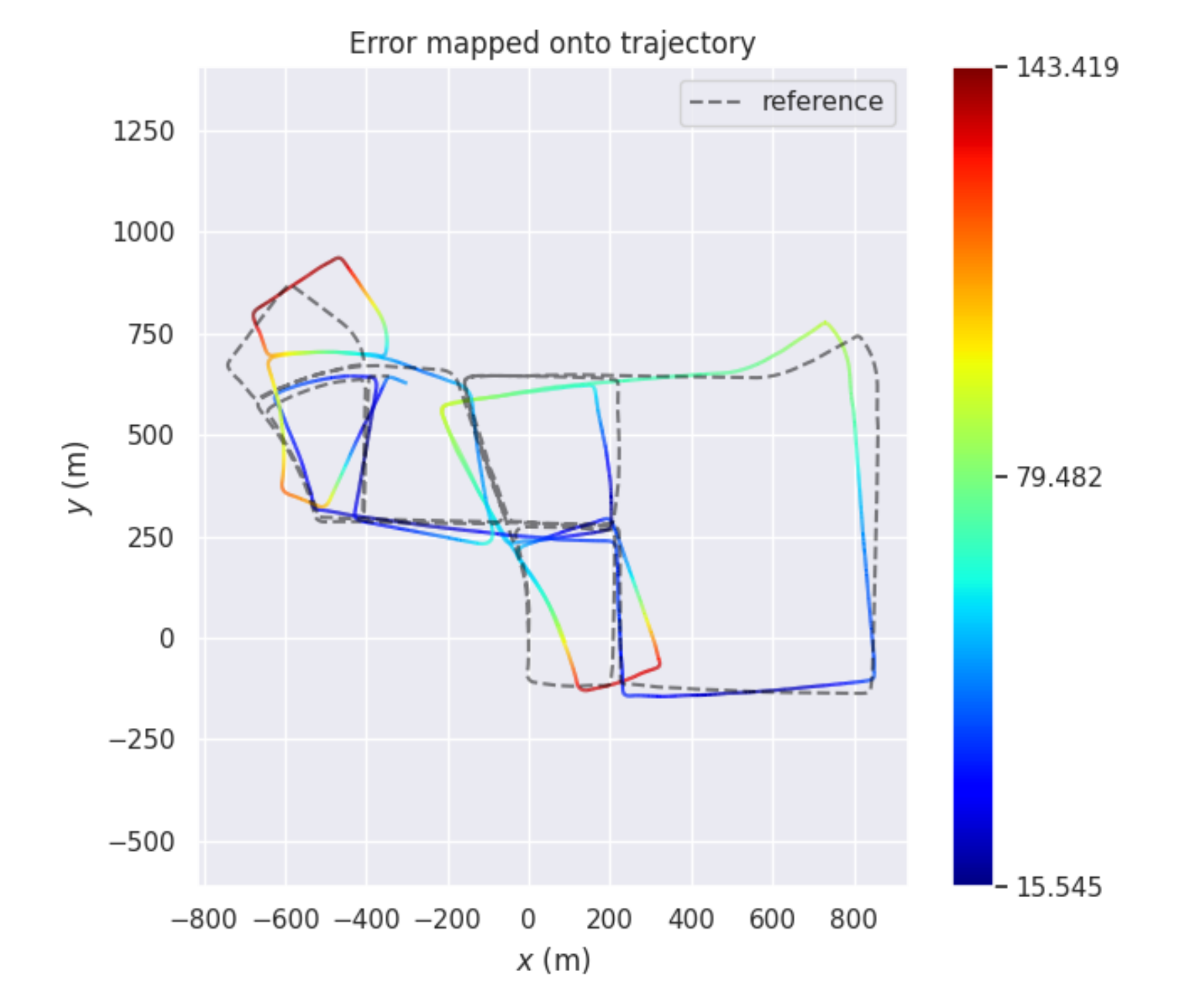}
  \end{minipage}%
  }%
  
  \centering
  \label{fig:kaist}
  \caption{ATE mapped onto Lvio-Fusion and VINS-Fusion's trajectory result of urban39-pankyo.}
\end{figure}

\subsection{Lvio-Fusion Assisted by TD3 agent}
\label{sec:experiment-rl}

To train the self-adaptive TD3 agent, we selected sequences 02, 05, 06, 07, 09 in KITTI dataset as the training sets and sequence 00 as the validation set. First, Lvio-Fusion extracted visual features and lidar features and calculated the IMU's pre-integration on all training sets. During the training stage, we divided the dataset into multiple parts with keyframes' size of 10 as environments and randomly selected one of them for TD3 agent training. Lvio-Fusion estimated poses with weights output by the agent, then, it returned a reward and the next observation. After finishing ten keyframes in current environment, a ``done" signal would be sent and selected another environment. 

\begin{figure} 
  \centering
  \includegraphics[width=8cm]{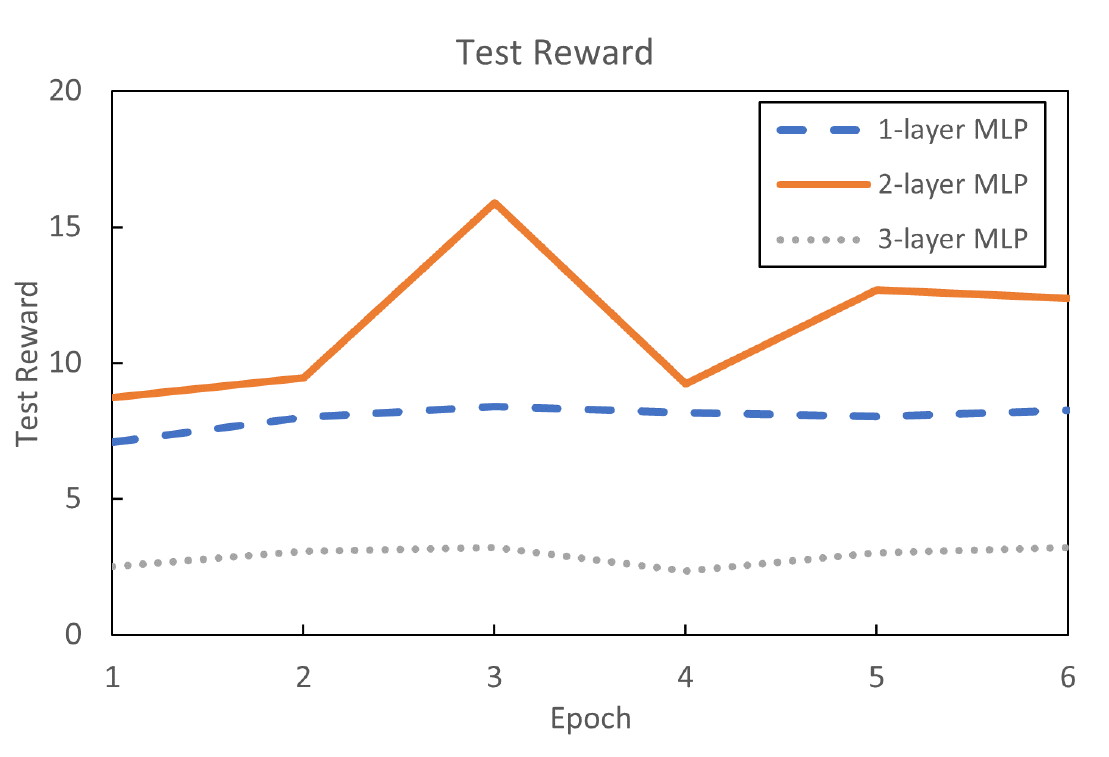}
  \caption{Test reward on each epoch.}
\label{fig:reward}
\end{figure}

In this experiment, we tried different depth of backbone networks. The Fig.\ref{fig:reward} shows test reward of each epoch. At the third epoch when test reward reached a peak, we ended training and saved the model for the next training of other training sets. Besides, different numbers of layers of MLP backbone for the agent were set up. By comparison, shown in Fig.\ref{fig:reward}, we selected two-layer MLP backbone with the highest reward and validated the agent.

In the validation phase, we turned off global optimization and used RPE to evaluate the results, which is more suitable for situations without global optimization. The results of Lvio-Fusion assisted by the self-adaptive algorithm on the KITTI dataset are shown in Table \ref{tab:rl}, and Fig.9 illustrates that self-adaptive algorithm significantly decreases error. And the average time-consuming of adjusting factors' weight of one keyframe is 5.56ms. Experiments have proved that the lightweight self-adaptive algorithm effectively improves the accuracy of the multi-sensor fusion.

\begin{table}[h]
  \centering
  \caption{RPE in KITTI dataset in meters}
  \label{tab:rl}
  \begin{tabular}{c|cc}
  \hline
  \multirow{2}{*}{Seq.} & Lvio-Fusion & self-adaptive Lvio-Fusion \\ \cline{2-3}
  & RMSE  & RMSE        \\ \cline{1-3}
  \rule{0pt}{10pt} 
  00 & 1.73    &  \textbf{1.39}   \\ \hline
  \end{tabular}%
\end{table}

\begin{figure}[h]
  \centering
  \subfigure[Lvio-Fusion]{
  \begin{minipage}[t]{1\linewidth}
  \centering
  \includegraphics[width=8cm]{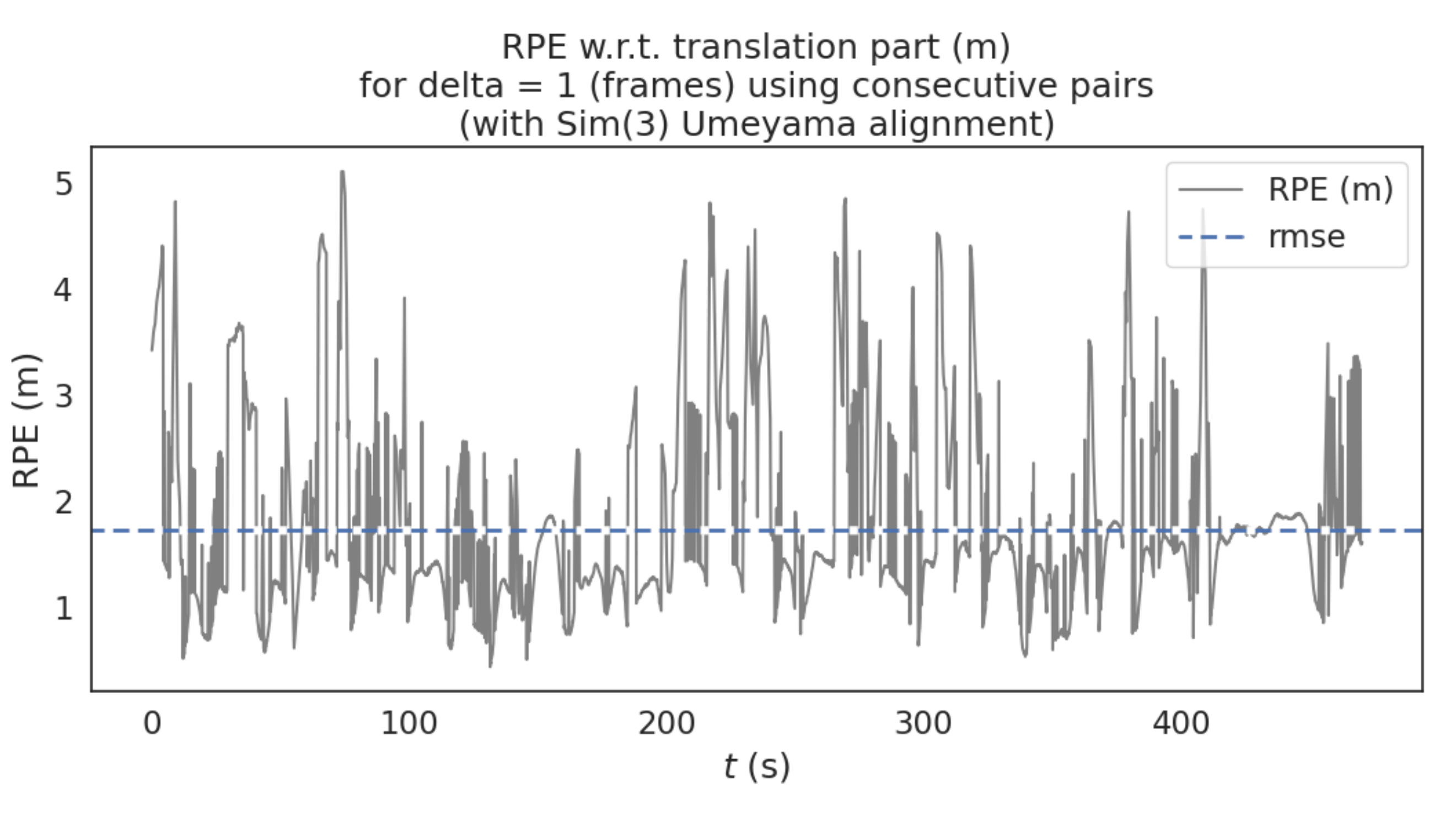}
  \end{minipage}%
  }%

  \subfigure[self-adaptive Lvio-Fusion]{
  \begin{minipage}[t]{1\linewidth}
  \centering
  \includegraphics[width=8cm]{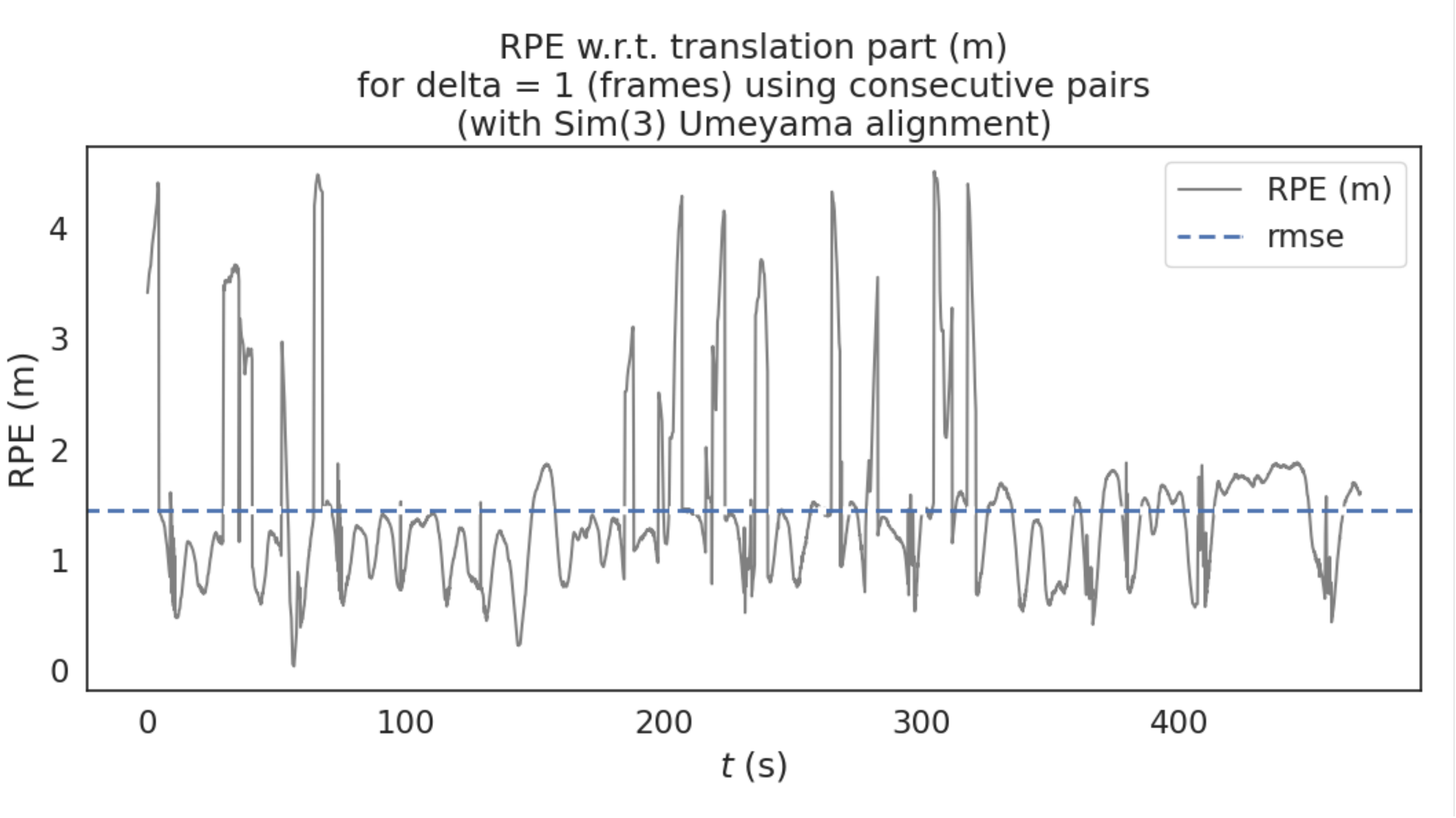}
  \end{minipage}%
  }%
  
  \centering
  \label{fig:self-adaptive}
  \caption{The Lvio-Fusion and self-adaptive Lvio-Fusion RPE in sequence 00.}
\end{figure}

\section{CONCLUSION}

We proposed Lvio-Fusion, a tightly coupled SLAM framework with stereo camera, lidar, IMU, and GPS, which fuses sensors based on factor graph optimization. Benefiting from this, other sensors can be easily fused into the framework as new factors. To eliminate accumulated drifts and provide global positioning, we design the segmented pose graph optimization with GPS constraint and loop closure constraint. To minimize the error caused by the fact that sensors have different performance in different environments, we used a lightweight deep reinforcement learning algorithm to adjust each factor's weight. The proposed method is thoroughly evaluated on the KITTI dataset and the KAIST urban dataset. The results show that Lvio-Fusion can achieve high accuracy, high robustness and real-time performance. But the self-adaptive algorithm need to be trained and evaluated with larger datasets for generalization. Also, we plan to conduct in-depth research on multimodal deep learning in the future for more information as input.


\bibliographystyle{IEEEtran}  
\bibliography{references}

\end{document}